\documentclass[letterpaper, 10 pt, conference]{ieeeconf}  

\IEEEoverridecommandlockouts                              

\usepackage{tabularx}
\usepackage{subfig}
\usepackage{orcidlink}
\usepackage{hyperref}

\usepackage{makecell}

\usepackage{array}
\newcolumntype{Z}{>{\centering\let\newline\\\arraybackslash\hspace{0pt}}X}

\usepackage{url}

\usepackage{array}
\usepackage{booktabs}
\usepackage{svg}

\usepackage{xargs}
\usepackage{multirow}

\usepackage{todonotes}

\newcommandx{\michael}[2][1=]{\todo[inline, linecolor=orange,backgroundcolor=orange!25,bordercolor=orange,author=Michael,#1]{#2}}
\newcommandx{\sleiman}[2][1=]{\todo[inline, linecolor=yellow,backgroundcolor=yellow!25,bordercolor=yellow,author=Sleiman,#1]{#2}}
\newcommandx{\christoph}[2][1=]{\todo[inline, linecolor=red,backgroundcolor=red!25,bordercolor=red,author=Christoph,#1]{#2}}

\newcommand\copyrighttext{%
	\footnotesize \copyright{ }2024 IEEE. Personal use of this material is permitted. Permission from IEEE must be obtained for all other uses, in any current or future media, including reprinting/republishing this material for advertising or promotional purposes, creating new collective works, for resale or redistribution to servers or lists, or reuse of any copyrighted component of this work in other works.}
\newcommand\copyrightnotice{%
	\begin{tikzpicture}[remember picture,overlay]
	\node[anchor=south,yshift=15pt,xshift=0pt] at (current page.south) {\parbox{\dimexpr\textwidth-\fboxsep-\fboxrule\relax}{\copyrighttext}};
	\end{tikzpicture}%
}

\title{\LARGE \bf
Causality-based Transfer of Driving Scenarios to Unseen Intersections
}

\author{Christoph Glasmacher\orcidlink{0000-0003-4826-9706}$^{1*}$, Michael Schuldes\orcidlink{0000-0003-2339-8157}$^{1*}$, Sleiman El Masri$^{2}$, Lutz Eckstein$^{1}$
\thanks{The work of this paper has been done in the context of the V4SAFETY project funded by the European Commission under grant agreement number $101075068$.
Views and opinions expressed, are those of the author(s) only and do not necessarily reflect those of the European Union or the European Climate, Infrastructure and Environment Executive Agency (CINEA). Neither the European Union nor the granting authority can be held responsible for them.}%
\thanks{Additionally, this research is funded by the VVM project research initiative, promoted by the German Federal Ministry for Economic Affairs and Climate Action (BMWK)}
\thanks{$^{*}$These authors contributed equally to this work and share first authorship.}
\thanks{$^{1}$Christoph Glasmacher, Michael Schuldes, and Lutz Eckstein are with Institute for Automotive Engineering RWTH Aachen University, Aachen, Germany {\tt\small firstname.lastname@ika.rwth-aachen.de}}%
\thanks{$^{2}$Sleiman El Masri is RWTH Aachen University, Aachen, Germany {\tt\small sleiman.el.masri@rwth-aachen.de}}%
}

\begin{document}

\maketitle
\thispagestyle{empty}
\pagestyle{empty}
\copyrightnotice
\vspace{-0.1in}

\begin{abstract}

Scenario-based testing of automated driving functions has become a promising method to reduce time and cost compared to real-world testing. 
In scenario-based testing automated functions are evaluated in a set of pre-defined scenarios.
These scenarios provide information about vehicle behaviors, environmental conditions, or road characteristics using parameters.
To create realistic scenarios, parameters and parameter dependencies have to be fitted utilizing real-world data. 
However, due to the large variety of intersections and movement constellations found in reality, data may not be available for certain scenarios. 
This paper proposes a methodology to systematically analyze relations between parameters of scenarios. Bayesian networks are utilized to analyze causal dependencies in order to decrease the amount of required data and to transfer causal patterns creating unseen scenarios.
Thereby, infrastructural influences on movement patterns are investigated to generate realistic scenarios on unobserved intersections.
For evaluation, scenarios and underlying parameters are extracted from the inD dataset.
Movement patterns are estimated, transferred and checked against recorded data from those initially unseen intersections.

\end{abstract}


\section{Introduction}
\label{sec:introduction}

In recent years, automated driving evolved rapidly.
With higher automation levels and handover of responsibilities, assuring and assessing the safety of these systems is key in automotive engineering.
To traditionally assure the safety of an automated driving function, theoretically billions of kilometers in field tests would be required which is too time-consuming and costly \cite{Winner2018a}. 
Therefore, scenario-based testing appears promising to reduce costs and accelerate the introduction of automated functions \cite{NEU20} as well as approximating its influence \cite{ROS20}.
Scenario-based testing entails assessing the automated driving functions based on a set of predefined scenarios in contrast to testing them on public roads where the events are random and specific conditions cannot be easily reproduced.
To test functions within scenario-based testing, a set of test scenarios has to be prepared according to an evaluation scope.
Recently, research focused on how to generate those scenarios data-driven or knowledge-based \cite{RIE20}.
A benefit of especially real-world data-driven approaches is, that they allow the derivation of the occurrence probability in the real-world under given recording biases.

Those generation methods often focus purely on dynamic objects at given specific intersections. 
Though, real-world is more complex and different intersections and variations have to be investigated.
However, many recent data acquisition methods focus on infrastructural fixed sensors.
Collecting data in this manner, lots of data is available within the observed constraints when having an automatized toolchain. 
To project conclusions on intersections beyond the observed traffic area, new data has to be collected. 
This is time-consuming and costly.

To decrease the need to acquire data from a lot of different intersections, this paper presents a methodology on how to estimate causal influences of intersection geometry on driving scenario parameters. 
Thus, a way to generate plausible trajectories on diverse and unseen urban intersections from scenario parameters is presented.
To validate the methodology, scenarios at intersections are extracted and distributions and dependencies are modeled from which several essential characteristics will be identified.
Bayesian networks are utilized to establish causal relationships between the scenario characteristics from which trajectories are reconstructed and transferred to other intersections. 
This will be evaluated on four different intersections comparing generated with real-world data using the inD dataset \cite{inDdataset}.

\section{Related Work}
\label{sec:related_work}

In current literature, methods to design scenarios and combine them with data are proposed. Thereby, scenarios are defined differently. Following, the general definition of a scenario as a sequence of scenes is used according to \cite{ISO34501}. 
Furthermore, according to \cite{VVM_D13}, a scenario can be described as an interpretable set of parameters which is called a logical scenario class. 
Scenarios with specific values for these parameters, set either from sampling from distributions or arbitrarily, are called concrete scenarios.
Following, the task to model scenario classes is described and split up into two steps: the parametrization of scenarios and the modeling of potential dependencies.

\subsection{Scenario parametrization}
\label{sec:parameter_space_exploration}

Multiple approaches have been developed to structure logical scenario classes.
\cite{SCH21} structures the environment in 6 layers to give guidance about general parts of a scenario.
\cite{AUTOontology22} builds on that by filling those layers with potentially relevant parameters based on knowledge. 
This method provides a valuable basis for a description of the scenarios defined by an evaluation scope, but does not determine the definition nor the importance of a parameter.

Focusing on scenarios and more detailed focusing on trajectories within, \cite{asano2011modeling} analyzes and models trajectory variations of left-turning vehicles at signalized intersections. 
For this, he uses a parametrization based on Euler spirals.
\cite{BEC22} uses Bezier curves to model trajectories for concrete scenarios.
\cite{WEB23} utilize cubic Hermite splines in a Frenet coordinate system based on OpenDRIVE to consider velocities additionally. 
Influences from the road network are thereby only implicitly considered utilizing the Frenet space.

\subsection{Modeling scenario parameter dependencies}
\label{sec:scenario_generation}

Based on a given parameter space, multiple methods have been developed that model dependencies between parameters while adding knowledge to reduce needed data.
The semantic information used within these approaches highly depends on the method. 
\cite{lotto2022modeling} proposes the use of copulas to capture the dependencies and correlations of stochastic parameters extracted from real-world data.
\cite{elrofai2018streetwise} uses kernel density estimations to achieve a similar goal. 
Both methods rely purely on correlations but do not include further semantic information.

Besides those pure data-driven methods, \cite{bagschik2018ontology} proposes a method to generate traffic scenarios using ontologies, which are knowledge-based systems that can be used to represent and reason about traffic concepts and relations. 
Although multiple scenarios can be generated, the methods lack a link to real-world distributions.
\cite{jesenski2019generation} proposes a method using a probabilistic graphical model to represent traffic to generate scenes, but lacks generating complete scenarios.
\cite{koopmann2022grasping} explores the causal relationships between the criticality influencing factors of real-world scenarios for automated driving systems based on Judea Pearl’s causal theory to describe scenarios, but lacks descriptions for the generation of scenarios. 
\cite{GLA23a} combines those approaches and uses data-driven probabilities in combination with constraints to generate realistic scenarios. 
However, how to come to those causal relations for infrastructural elements systematically and how to extrapolate movement patterns to other unseen intersections is not described.

\section{Methodology}
\label{sec:methodology}

Projecting driving scenarios from one intersection to another is not trivial since the infrastructure characteristics influence the distributions of scenario parameters as well as external and unknown factors.
So, different driver behaviors can be observed when changing the infrastructure or even when approaching an intersection from a different arm.
To be able to map scenarios or trajectories to different intersections, relevant infrastructure elements have to be taken into account as causal factors influencing driving behavior. 
Respectively, this change of infrastructure should be reflected in parameter distributions describing the scenario.

\begin{figure*}[tb]
	\centering
	\vspace{0.06in}
	\includegraphics[width=\linewidth]{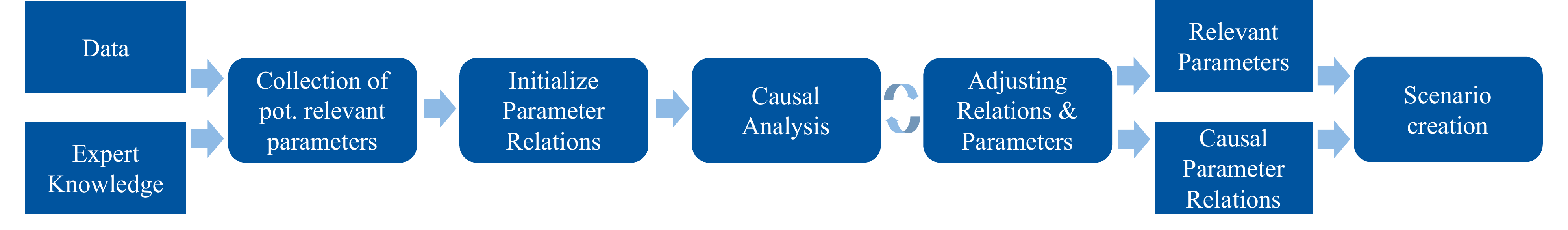}
	\caption{Methodology for scenario setup}
	\label{fig:methodology}
\end{figure*}

To account for those influences, elements of the 6 layer model are abstracted into interpretable parameters as the representation in a multidimensional parameter space utilizing causal analysis (see Fig.~\ref{fig:methodology}).
Thereby, the parameter space is not interpreted as a collection of distinct parameters, but as a set of parameters and their relevant relations.
As the first step, we identify potential relevant parameters based on the objective of the scenario, expert knowledge, and available data.
Then, we model the relation and interaction of those parameters in a causal Bayesian network, by utilizing data from concrete scenarios.
Finally, a causal analysis is performed on the network to adjust relevant parameters and their relations. 
From this space, scenarios can again be sampled and translated into executable scenarios with optional variations of infrastructural parameter values taking over movement patterns from seen intersections.

Within this process, two kinds of parameters are distinguished within a scenario to describe patterns and relations sufficiently: direct and indirect parameters:
\begin{itemize}
	\item \textbf{Direct parameters} are those directly serving for constructing executable scenarios. They are translated into a description interpretable by a simulation tool. 
	\item \textbf{Indirect parameters} are not directly used for translation into description understandable by a simulation tool, but they are used to make relations within the parameter space explicit. So, they have an indirect impact on the resulting scenario. E.g., if a road curvature is not modeled explicit, it could be an indirect parameter if it has an impact on the path of a road user.
\end{itemize}

\subsection{Initial Parameter and Relation Setup}
\label{sec:initialize_parameter_relations}

To model a scenario, an initial set of parameters is derived. This is done systematically for each of the six layers utilizing the ontology A.U.T.O \cite{AUTOontology22}, an underlying scenario concept \cite{WEB23} and available information from given data sources \cite{OMEGAformat}.
According to the definition of direct and indirect parameters, these parameters can be velocities as well as the occurrence of a pedestrian crosswalk although that may not be modeled in a simulation.
These parameters have to be set in relation to each other. 
To model the relations, a causal Bayesian network is used.
The known relations from the ontology and scenario concept are taken over into the structure of the Bayesian network as edges.

Based on this knowledge-based determination, the set of relations is scrutinized. Additional edges may be included by using optimization approaches applied to real-world data or existing relations may be deleted. 
To assess relations, the Greedy Hill Climbing search algorithm \cite{greedyhillcliming} is used.
The algorithms identify relations between the parameters $G$ using correlations from the data $D$ to derive relevant dependencies from correlation probabilities ($P$) assessing the likelihood score ($S$) (see Eq.~\ref{eq:likelyhood_score}, Eq.~\ref{eq:probability_DAG}). 
In order to not start from scratch with the learned approach, already known relations are set as necessary constraints. 

\begin{equation}
S(G:D) = LL(G:D) - \phi(|D|) \cdot ||G||
\label{eq:likelyhood_score}
\end{equation}
\begin{equation}
P(D|G) = \int P(D|G, \Theta_G)\cdot P(\Theta_G | G) \cdot
d\Theta_G
\label{eq:probability_DAG}
\end{equation}
The combined use of knowledge-based and data-driven methods leads to a better quality because in large parameter spaces, a systematic analysis of the edges is crucial.

\subsection{Causal Analysis}
\label{sec:causal_analysis}

After having an initial set of parameter relations established, these relations are scrutinized using a causality analysis.
Within the causality analysis, relations between parameters are investigated qualitatively and quantitatively.
For both, the influence of a parameter on the network is investigated under the constraints of representing causal effects in the scenario.
This is different from the used Greedy Hill Climbing algorithm since this focuses on correlation, but not causality. 
In contrast to correlation, causality makes a distinction between cause and effect.
So, the direct and explicit effect of a variable on another is analyzed.
For the qualitative analysis, expert knowledge, available literature, and physical effects are used.

The do-calculus is used for the quantitative analysis. 
The do-operator assesses the controlled direct effect from one parameter $X$ with a given state $x$ to another parameter $Y$ while considering all other possible variables $i$ which are not intervened (see Eq.~\ref{eq:do_calculus}). 
The calculation thereby gives a clear indication of the causal dependency.
\begin{equation}
P(Y | do(X=x)) = \sum_{i} P(Y | X=x, i) \cdot P(i)
\label{eq:do_calculus}
\end{equation} 

After causal investigation on effects, causal relations are adjusted according to the outcome of the analysis. 
Outcomes from both, qualitative and quantitative analysis are used.
Depending on the data, it can be justifiable to include edges even in cases where data does not give a hint towards a causal relation if a clear effect is obvious from physics or given in literature. 
Reasons could be physical constraints, known effects, or known bias in data which should be overcome.
E.g., a correlation between weather and vehicle velocities is reasonable, but may not be found in data due to the lack of diverse weather conditions.
If both, quantitative do-calculus and qualitative logic do not hold for a set relation, it is removed.
In an iterative process, all potential relations are checked to define a set of valid causal parameter relations.

\begin{figure}[tb]
	\centering
	\vspace{0.06in}
	\includegraphics[width=\linewidth]{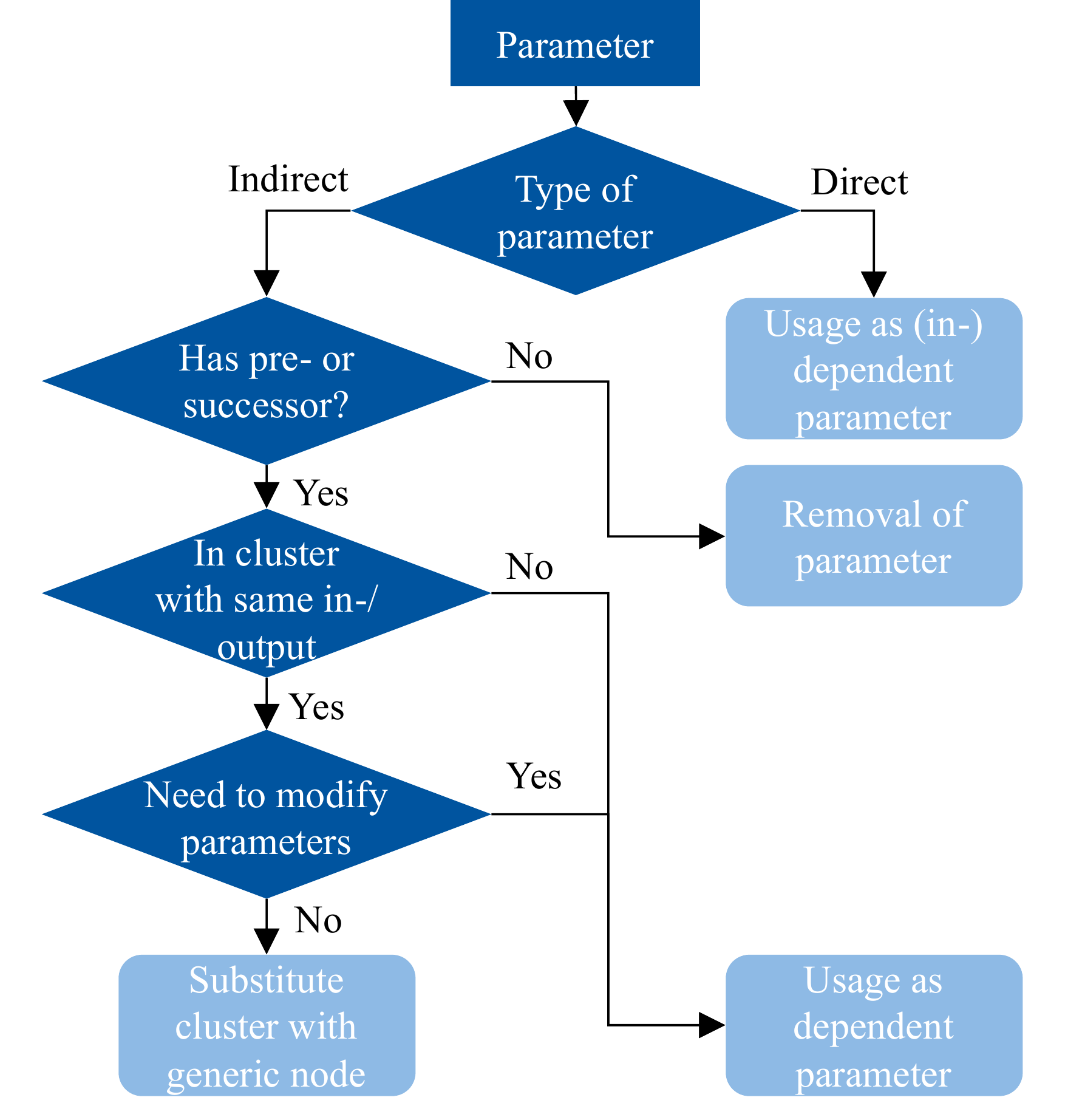}
	\caption{Flowchart of parameter handling}
	\label{fig:flowchart_parameterhandling}
\end{figure}

After completion of the causal analysis, the remaining parameter connections are investigated. 
Thereby, a distinction between direct and indirect parameters (see Fig.~\ref{fig:flowchart_parameterhandling}) has to be made. 
Since direct parameters are needed for scenario creation, they can not be neglected.
So, if there is no relation to known effects, the parameter is assumed to be independent from other investigated parameters.
To simplify the network, indirect parameters can be deleted if there is no direct causal relation to another parameter. 
Since the aim is to describe relations, they are not useful within the framework if they do not have causal effects on other parameters.
This also holds true for clusters of indirect parameters. 
If a cluster of indirect parameters is disconnected from direct parameters, those can be removed. 
In the case that a cluster of indirect parameters only has one in- or output relation, it can be substituted by a general node representing all indirect parameters unless those parameters should be individually varied within scenario variations.

After simplifying the network through the removal and substitution of nodes, the given network describes all relevant parameters and relevant causal relations of the scenario.

\subsection{Scenario Generation}
\label{sec:method_scenario_generation}

The created causal Bayesian network (Sec.~\ref{sec:causal_analysis}) is then used to generate scenarios.
Thereby, real-world data is included in the causal Bayesian network to estimate the conditional probabilities similarly as in \cite{GLA23a}.
Importantly, data has to be regarded not for all intersections, but for a diverse set to be able to assess conditional probabilities for infrastructural influences.
The fitted network is used to generate parameter sets for unseen intersections, by setting the infrastructural conditions according to the characteristics of the desired intersection.
Since the fitted model inhibits the estimated causal effects, the sampled parameters are in accordance with those causalities and therefore plausible.
Finally, the sampled parameter set can be translated into a simulation executable format.

\section{Results}
\label{sec:results}

\begin{figure}[tb]
	\centering
	\subfloat[Frankenburg]{
		\includegraphics[width=0.44\linewidth]{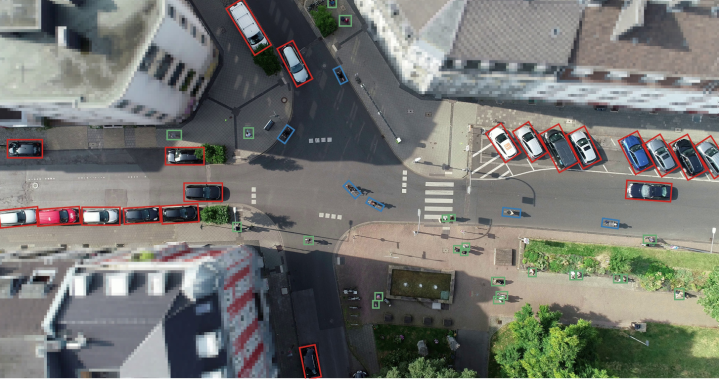}
	}
	\hspace{0.005\linewidth}
	\subfloat[Neukölln]{
		\includegraphics[width=0.44\linewidth]{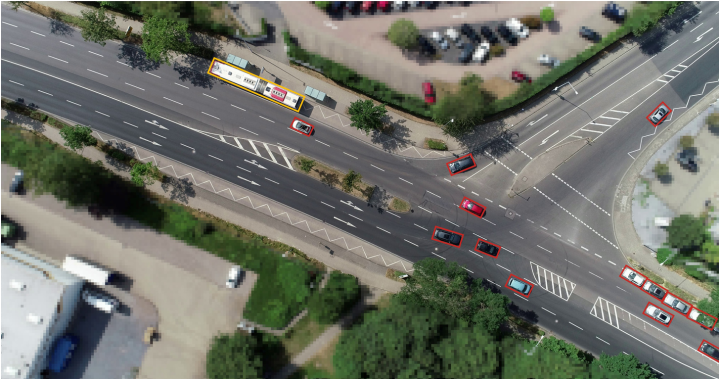}
	}
	\caption{Exemplary intersections from inD dataset \cite{inDdataset}}
	\label{fig:inD_intersections}
\end{figure}

In the following, the proposed methodology is applied to generate plausible trajectories on unobserved intersections. 
Thereby, it is distinguished between left turns, right turns, and passing straight maneuvers within different scenarios. 
As a data source, the inD dataset is utilized which contains 13.499 trajectories on four different intersections, two of these are shown in Fig.~\ref{fig:inD_intersections}

The proposed methodology is evaluated in three steps.
Firstly, it is tested if a derived scenario representation can be used to recreate individual trajectories, and the loss due to parametrization is evaluated.
Afterward, we apply the causal analysis and use the trained model to transfer trajectories to unseen intersections, where we compare the generated trajectories to real-world observations.
Finally, the influence of specific characteristics is investigated through the example of a construction site.

\subsection{Scenario representation}

\begin{figure}[tb]
	\centering
	\vspace{0.06in}
	\includegraphics[width=\linewidth]{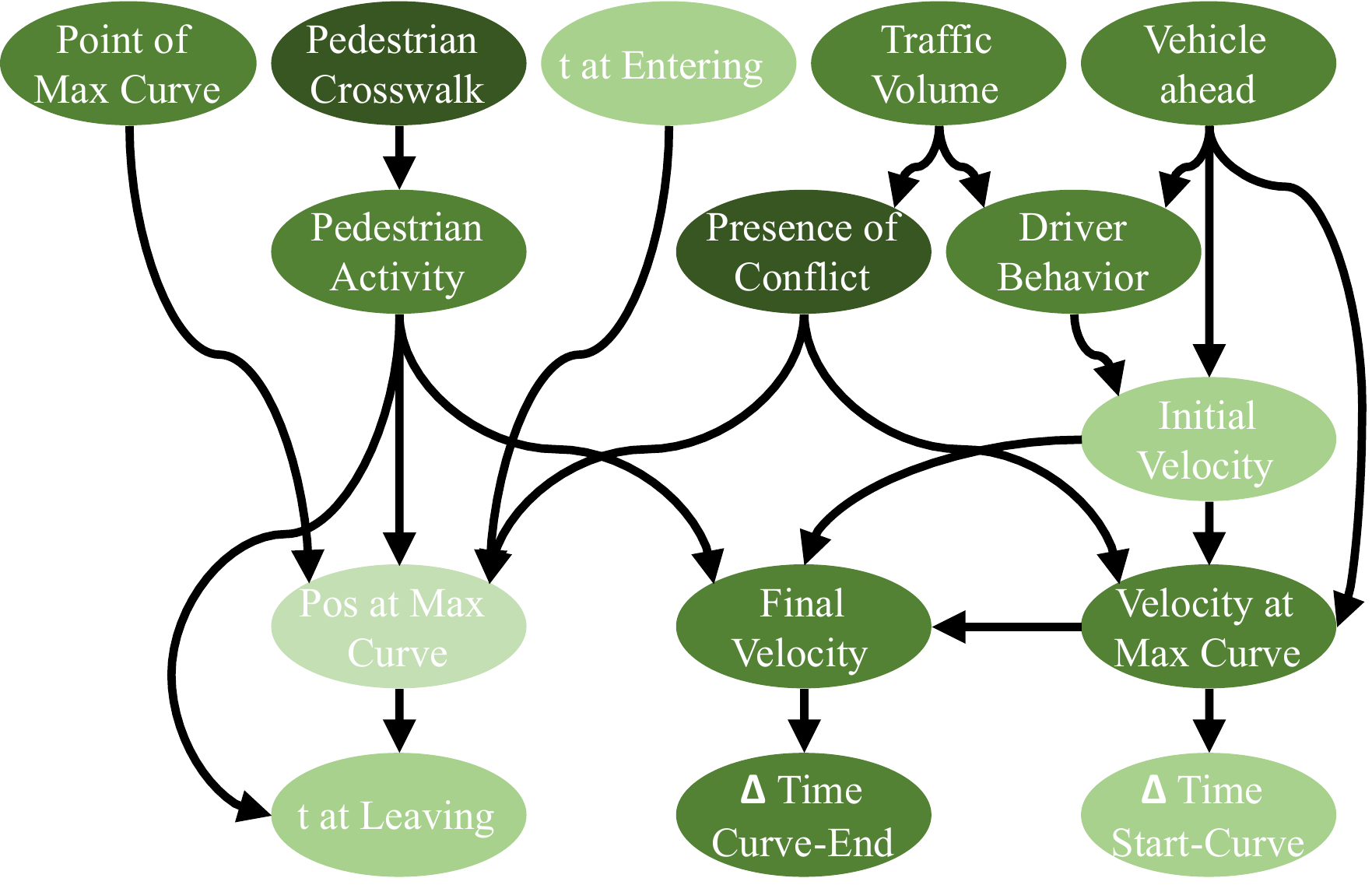}
	\caption{Complete causal Bayesian network. The darker the node, the less equally distributed are the occurring parameter values.}
	\label{fig:bayesian_network}
\end{figure}

To represent trajectories on intersections, the spline-based method from \cite{WEB23} along with the road-based Frenet coordinate system is used.
The traveled distance on a lane is described by the coordinate $s$ and the deviation from the lane centerline through the coordinate $t$.
$t$ is oriented so that through increasing $t$ the object move further to the left of the centerline.
Furthermore, infrastructural and scenario-defining parameters are added resulting in a total number of 15 parameters (see Fig.~\ref{fig:bayesian_network}). 
The number of parameters could be reduced when applying only annotated data for a scenario category so that e.g., the presence of conflict is implicitly covered by the chosen scenario category and underlying fitted data.

These parameters are extracted for individual maneuvers and intersection arms from the inD dataset.
Comparing the trajectories reconstructed from the parameters to the real-world trajectories, it can be seen, that they are relatively similar (see Table~\ref{tab:trajectory_parametrization}). 
Errors occur due to the abstraction made within the pure parametrization.
Additionally, the discretization in the Bayesian network has a negative effect on the reconstruction performance.
Anyhow, those steps are necessary to fit distributions due to the limited amount of data available.
The errors in the dynamic time warping metric (DTW) are larger compared to the Fréchet distance due to stops of vehicles at intersections.
Since the spline modeling uses the time as a parameter and therefore models time and position at the same time, still stand of an object can not be modelled in sufficient detail. 

\begin{table}[!ht]
	\centering
	\caption{Average reconstruction loss of trajectories due to parametrization with considering a discretization step for the Bayesian network (\textit{discrete}) and without this discretization (\textit{raw}).}
	\label{tab:trajectory_parametrization}
	\begin{tabular}{|l|l|ll|ll|}
		\hline
		\multirow{2}{*}{\textbf{Intersection}} & \multirow{2}{*}{\textbf{Turn}} & \multicolumn{2}{l|}{\textbf{Fréchet}} & \multicolumn{2}{l|}{\textbf{DTW}}     \\ \cline{3-6} 
		&                                & \multicolumn{1}{l|}{raw}   & disc. & \multicolumn{1}{l|}{raw}   & disc. \\ \hline
		\hline
		Frankenburg                            & N-E                             & \multicolumn{1}{l|}{0.56}  & 1.05     & \multicolumn{1}{l|}{5.58}  & 11.78    \\ \hline
		Frankenburg                            & S-E                              & \multicolumn{1}{l|}{0.31}  & 0.86     & \multicolumn{1}{l|}{2.75}  & 9.46     \\ \hline
		Neukölln                               & N-E/W-N                          & \multicolumn{1}{l|}{0.83}  & 1.52     & \multicolumn{1}{l|}{11.68} & 20.28    \\ \hline
	\end{tabular}
\end{table}

\begin{figure}[tb]
	\centering
	\vspace{0.06in}
	\includegraphics[width=\linewidth]{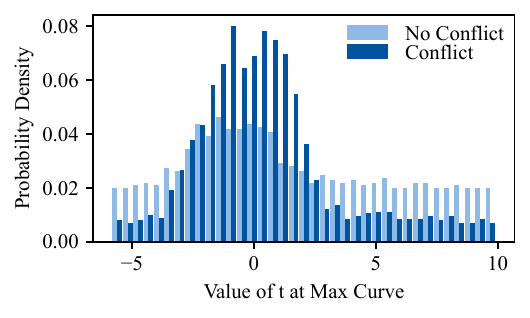}
	\caption{Causal effect of Conflict on deviation from lane centerline}
	\label{fig:t_conflict}
\end{figure}

Within the used parameter representation, different effects are taken into account.
Especially those focusing on infrastructural elements, dynamic elements, and constellations are assessed. 
Both, layer 1 and layer 4 parameters of the 6 layer model \cite{SCH21} are interconnected so that both have to be considered to assess relevant influences for trajectories. 
E.g., the presence of a conflict with another road user has a significant impact on the trajectory of a vehicle (see Fig.~\ref{fig:t_conflict}).
It can be seen that road users drive more strictly in the middle of the road when other road users are present compared to the absence of potential conflicts.
So, to reliably attribute causes, multiple aspects have to be considered at the same time.

Similarly, when applying do-calculus, it can be seen that the velocity of a vehicle is influenced by the presence of a conflict with other road users, as they approach intersections more slowly in these cases.

\subsection{Transfer to unseen intersections}

To investigate the influence of infrastructural elements, it is evaluated if behavior can be transferred to different intersections without including trajectory data of those new intersections or of different turns of the same intersection.
For the evaluation, not individual trajectories are compared to each other, but their distributions are compared. 
This is done since the Bayesian network models the relations and causalities statistically and not as a one-to-one mapping. 
A transfer of a single trajectory would require the classification of the behaviors and influences of single individuals, which is considered out of scope. 

\begin{figure}[tb]
	\centering
	\subfloat[Real-world]{
		\includegraphics[width=0.44\linewidth]{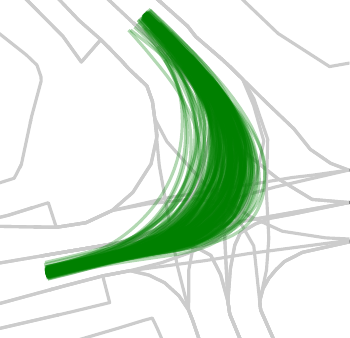}
	}
	\hspace{0.005\linewidth}
	\subfloat[Generated]{
		\includegraphics[width=0.44\linewidth]{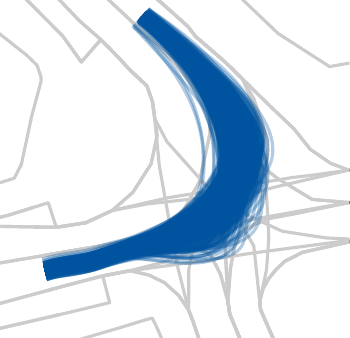}
	}
	\caption{Reconstruction of left turn trajectories on the Bendplatz intersection of the inD dataset.}
	\label{fig:trajectories_on_bendplatz}
\end{figure}

The Bayesian network is fitted based on a set of trajectories and intersections not including the intersection to assess. 
To evaluate the performance, the collection of generated trajectories is compared with real-world recorded trajectories on the given intersection and turn.
This is necessary since the maneuvers of different road users may vary due to different characteristics in the driving behavior of the traffic participants.
In particular The Bayesian network is trained on the north-east turn of the Frankenburg intersection and the north-east and west-north intersection of the  Neuköllnstraße intersection. 
The model is then used to sample new trajectories for the west-north left turn of the Bendplatz intersection. 
Fig.~\ref{fig:trajectories_on_bendplatz} shows these generated sampled trajectories and compares them to the real-world trajectories found in the inD dataset.
It can be observed, that the method is able to generate plausible trajectories, though the spread of the trajectories in the apex of the turn is smaller. 

\begin{figure}[tb]
	\centering
	\subfloat[at start of turn ($s=5m$)]{
		\includegraphics[width=0.44\linewidth]{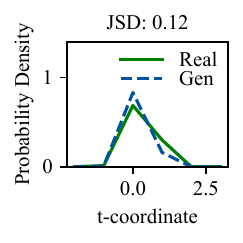}
	}
	\hspace{0.005\linewidth}
	\subfloat[at end of turn ($s=30m$)]{
		\includegraphics[width=0.44\linewidth]{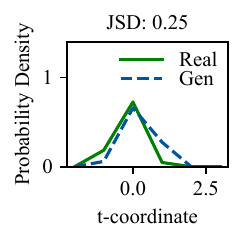}
	}
	\caption{Jensen Shannon divergence (JSD) of parameter t of generated compared to real.}
	\label{fig:js_distributions}
\end{figure}

To quantitatively assess the performance, the Jensen-Shannon divergence \cite{JSD97} applied to the distribution of trajectories is used. 
To simplify the non-trivial calculation of distribution of trajectories, the distribution of the $t$-coordinate (deviation from lane centerline) is computed for different points on the path of the turn.
This can be seen as evaluating the cross-section of the trajectories at different points.
It can be seen that the set of generated and original trajectories are similarly distributed, but a slight deviation can be observed (see Fig.~\ref{fig:js_distributions}). 
This deviation can be due to missing influencing factors and low statistical significance due to the limited amount of input data.
Additionally, the errors due to parametrization described above could play a role.

\subsection{Influence of special infrastructural characteristics}

\begin{figure}[tb]
	\centering
	\vspace{0.06in}
	\includegraphics[width=\linewidth]{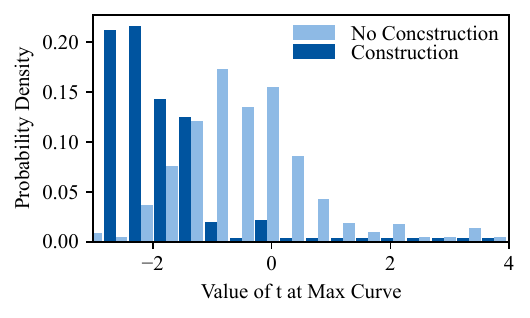}
	\caption{Causal effect of construction site on deviation from lane centerline}
	\label{fig:construction_site}
\end{figure}

Besides the mapping to different intersections, it is investigated how the method copes with small sample sizes.
Therefore, the influence of a construction site on the observed trajectories is evaluated with the given method, since those are rare in the dataset.
The framework allows to quantify the influence on the behavior in these situations and to derive causal relations.
Using the do-calculus and investigating the effect from a far-side construction site it can be seen that it influences the distribution of the other parameters significantly (see Fig.~\ref{fig:construction_site}).
Whereas vehicles drive on average around the center line of their lane if no construction site is present, the presence of such a construction site causes a shift away from the center lane, away from the construction site.
Furthermore, the distribution has a smaller spread, indicating that vehicles drive relatively similar.
Thus, such influences and parameters have to be considered in modeling, although the sample size is rather small. 
This result demonstrates that such influences can be modeled with the proposed method.

\section{Discussion}
\label{sec:discussion}
The results shown that the method can fulfill the minimum requirement of recreating the training data. 
The chosen parametrization despite not being lossless, introduces an acceptable amount of error.
This enables the investigation of causalities and of the transfer of trajectories to different intersections.
Secondly, it is shown how the dependencies between parameters can be analyzed, and how the effects can be investigated using causal Bayesian networks. 
The following is important to mention: 
Effects between parameters can be non-causal and just correlated. 
To state causality, the found effects have to be investigated and argued for.
The proposed method can be extended to complete scenarios since it abstracts dynamics to a general parametrization.

Furthermore, it is shown that a transfer to different intersections knowing infrastructural attributes is feasible.
This holds true as long as the underlying causalities in the new intersections are similar to the ones found in training data.
The degree of required similarity thereby depends on the used bin size of the Bayesian network. 
If the bin size is increased, fewer data is needed for sufficient confidence in conditional probability distributions. 
But, if it is increased too much, trajectories become inaccurate due to discretization, and therefore less realistic.
This trade-off always has to be considered.

Required data can be further decreased using constraints as described in \cite{GLA23a} utilizing additional knowledge besides probabilistic causal relations.

\section{Conclusion}
\label{sec:conclusion}

This paper proposed a methodology to model parameter relations and movement patterns and applies them to unseen scenarios focusing on the projection to unseen intersections.
Causal analysis in combination with a learning-based Bayesian network algorithm is used to identify relevant parameters and parameter dependencies.
This method is evaluated by extracting scenarios from naturalistic real-world data and generating realistic scenarios from this data.
Thereby, a reproducibility of real-world scenarios as well as a similar representativeness is shown.
This holds true for seen and previously unseen intersections so that movement patterns are described sufficiently well and scenario generation on unobserved intersections is viable.
Furthermore, influences on movement patterns, such as those of construction sites, are investigated, but more data is required to draw conclusions on general parameter dependencies.

\addtolength{\textheight}{-12cm}   

\bibliographystyle{IEEEtran}
\bibliography{bibtex/bib/IEEEabrv, bibtex/bib/paper}

\end{document}